\documentclass[tog]{acmsiggraph}
\usepackage{times}
\usepackage{epsfig}
\usepackage{graphicx}
\usepackage{tabularx}
\usepackage{multirow}
\usepackage{colortbl}
\usepackage{subfig}
\usepackage{amsmath}
\usepackage{algorithm}
\usepackage{appendix}
\usepackage{arydshln}
\usepackage[noend]{algpseudocode}
\usepackage{hhline}
\usepackage{amssymb}
\usepackage[percent]{overpic}
\usepackage[usenames,dvipsnames,svgnames,table,xcdraw]{xcolor}

\TOGonlineid{0301}
\TOGvolume{0}
\TOGnumber{0}
\TOGarticleDOI{1111111.2222222}
\TOGprojectURL{}
\TOGvideoURL{}
\TOGdataURL{}
\TOGcodeURL{}
\newcommand{\dbname}{ Florentine }
\newcommand{\tdash}{\cdashline{1-4}[0.5pt/0.5pt]}

\newcommand{\comment}[1]{}
\newcommand{\mat}[1]{\mathbf{#1}}

\renewcommand{\algorithmicrequire}{\textbf{Input:}}
\renewcommand{\algorithmicensure}{\textbf{Output:}}

\newcommand{\colora}{CAFEB8}
\newcommand{\colorb}{FDFD96}
\newcommand{\colorc}{63E9FC}
\definecolor{fmcolor}{HTML}{FDFD96}
\definecolor{cmcolor}{HTML}{63E9FC}
\definecolor{hcolor}{HTML}{CAFEB8}

\title{Deep video gesture recognition using illumination invariants}

\author{Otkrist Gupta
\and
Dan Raviv\\Massachusetts Institute of Technology
\and
Ramesh Raskar}
\pdfauthor{Otkrist Gupta}


\begin{document}

\teaser{
  \begin{center}
  \includegraphics[width=1\textwidth,height=2in]{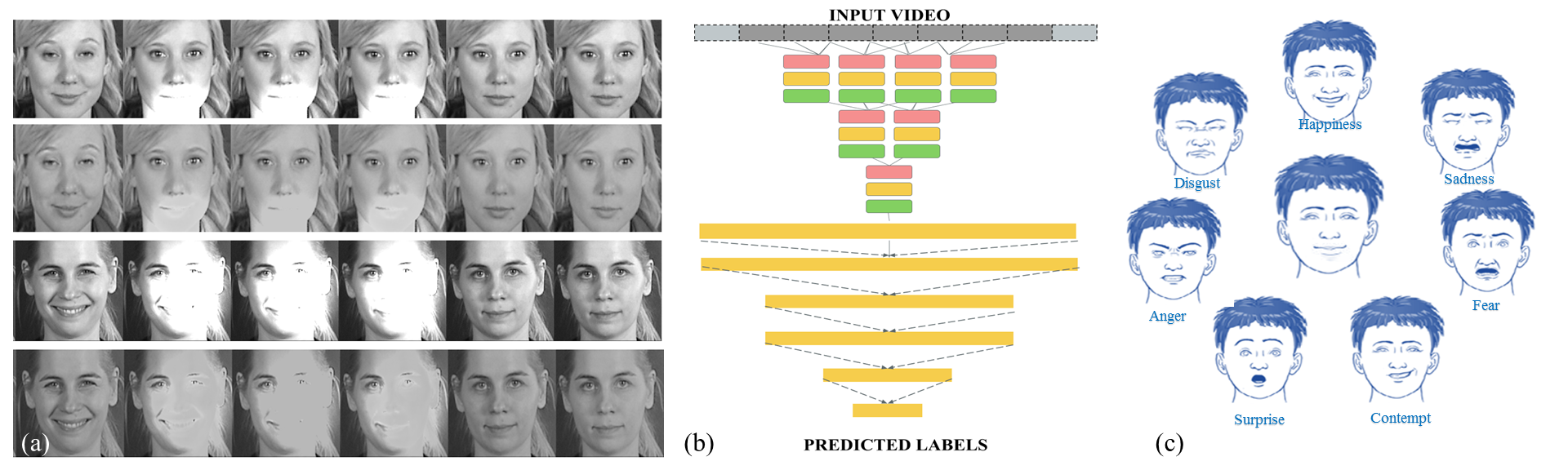}
  \caption{Automated facial gesture recognition is a fundamental problem in human computer interaction. While tackling real world tasks of expression recognition sudden changes in illumination from multiple sources can be expected. We show how to build a robust system to detect human emotions while showing invariance to illumination.}
  \end{center}
}

\maketitle

\begin{abstract}
In this paper we present  architectures based on deep neural nets for gesture recognition in videos, which are invariant to local scaling. We amalgamate autoencoder and predictor architectures using an adaptive weighting scheme coping with a reduced size labeled dataset, while enriching our models from enormous unlabeled sets. We further improve robustness to lighting conditions by introducing a new adaptive filer based on temporal local scale normalization.
We provide superior results over known methods, including recent reported approaches based on neural nets.



\end{abstract}

\keywordlist





\section{Introduction}

Human beings, as social animals, rely on a vast array of methods to communicate with each other in the society. Non-verbal communication, that includes body language and gestures, is an essential aspect of interpersonal communication. In fact, studies have shown that non-verbal communication accounts for more than half of all societal interactions \cite{frith2009role}. Studying facial gestures is therefore of vital importance in fields like sociology, psychology and automated recognition of gestures can be applied towards creating more user affable software and user agents in these fields.

Automatic gesture recognition has wide implications in the field of human computer interaction. As technology progresses, we spend large amounts of our time looking at screens, interacting with computers and mobile phones. In spite of their wide usage, majority of software interfaces are still non-verbal, impersonal, primitive and terse. Adding emotion recognition and tailoring responses towards users emotional state can help improve human computer interaction drastically \cite{cowie2001emotion,zhang2015learning} and help keep users engaged. Such technologies can then be applied towards improvement of workplace productivity, education and telemedicine \cite{kolakowska2014emotion}. Last two decades have seen some innovation in this area \cite{klein1999computer,cerezo2007interactive,andre2000exploiting} such as humanoid robots for example \textit{Pepper} which can both understand and mimic human emotions.


\begin{figure*}[t]
\begin{center}
\includegraphics[width=\linewidth]{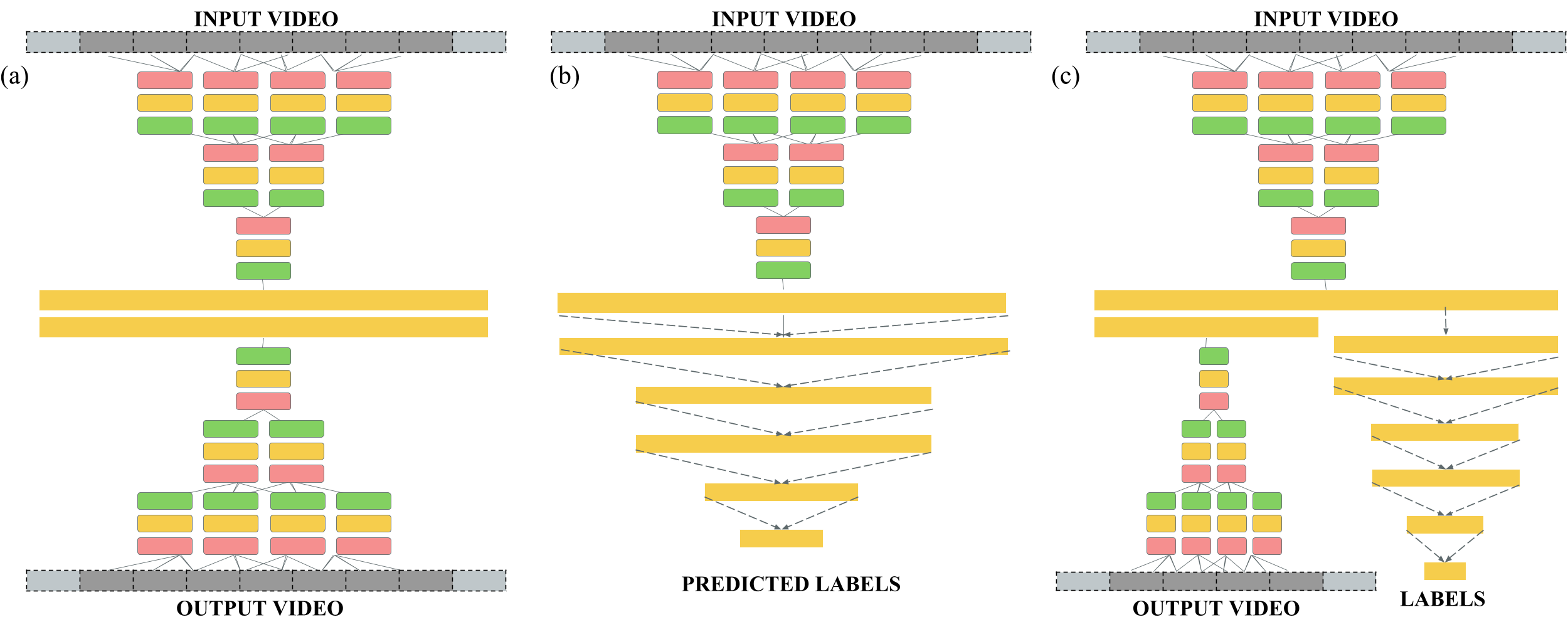}
\end{center}
   \caption{Schematic representation of deep neural networks for supervised and unsupervised learning. We use pink boxes to denote convolutional layers, yellow boxes denote \textit{rectified linear unit} layers and green boxes indicate normalization layers. Our technique combines unsupervised learning approaches (a) with labeled prediction (b) to predict gestures using massive amounts of unlabeled data and few labeled samples. Autoencoder (a) is used to initialize weights and then predictor (b) is fine tuned to predict labels.}
\label{fig:semisupervisedlearner}
\end{figure*}

Modeling and parameterizing human faces is one of the most fundamental problems in computer graphics \cite{liu2014deeply}. Understanding and classification of gestures from videos can have applications towards better modeling of human faces in computer graphics and human computer interaction. Accurate characterization of face geometry and muscle motion can be used for both expression identification and synthesis \cite{pighin2006synthesizing,wang2013capturing} with applications towards computer animation \cite{cassell1994animated}. Such approaches combine very high dimensional facial features from facial topology and compress them to lower dimensions using a series of parameters or transformations \cite{waters1987muscle,pyun2003example}. This paper demonstrates how to use deep neural networks to reduce dimensionality of high information facial videos and recover the embedded temporal and spatial information by utilizing a series of stacked autoencoders.

Over the past decade algorithms for training neural nets have dramatically evolved, allowing us to efficiently train deep neural nets \cite{hinton2006fast,jung2015joint}. Such models have become a strong driving force in modern computer vision and excel at object classification \cite{krizhevsky2012imagenet}, segmentation and facial recognition \cite{taigman2014deepface}. In this paper we apply deep neural nets for recognizing and classifying facial gestures, while pushing forward several architectures.  We obtain high level information in both space and time by implementing $4\mathbb{D}$ convolutional layers and training an autoencoder on videos.
Most of neural net applications use still images as input and rely on convolutional architectures for automatically learning semantic information in spatial domain. Second, we reface an old challenge in learning theory, where not all datasets are labeled.
Known as semi-supervised learning, this problem, once again, attracts attention as deep nets require massive datasets to outperform other architectures. Finally, we provide details of a new normalization layer, which robustly handles temporal lighting changes within the network itself. This new architecture is adaptively fine tuned as part of the learning process, and outperforms all other reported techniques for the tested datasets.
We summarize our contributions as follows:

\subsection{Contributions}

\begin{enumerate}

\item We develop a scale invariant  architecture for generating illumination invariant deep motion features.

\item  We report state of the art results for video gesture recognition using spatio-temporal convolutional neural networks.

\item We introduce an improved topology and protocol for semi-supervised learning, where the number of labeled data points is only a fraction of the entire dataset.


\end{enumerate}

\section{Related Work}

Machine learning strategies such as random forests or SVMs combined with local binary features (or sometimes facial fiducial points) have been used for facial expression recognition in the past \cite{kotsia2007facial,michel2003real,shan2005robust,dhall2011emotion,walecki2015variable,presti2015using,vieriu2015facial}. Other intriguing methodologies include performing emotion recognition through speech \cite{nwe2003speech,schuller2004speech}, using temporal features and manifold learning \cite{liu2014learning,wang2013capturing,kahou2015emonets,chen20153d} and combining multiple kernel based approaches \cite{liu2014combining,senechal2015facial}. Majority of such systems involve a pipeline with multiple stages - face discovery, face alignment, feature extraction and landmark localization followed by classification of labels as the final step. Our approach combines all of these phases (after face detection) into the neural net which takes entire video clip as input.

Recently, deep neural nets have triumphed over traditional vision algorithms, thereby dominating the world of computer vision. Deep neural networks have proven to be an effective tool to classify and segment high dimensional data such as images \cite{krizhevsky2012imagenet,szegedy2015going}, audio and videos \cite{karpathy2014large,tran2014learning}. With advances in convolutional neural nets, we have seen neural nets applied for face detection \cite{taigman2014deepface,zhao2015joint} and expression recognition \cite{abidin2012neural,gargesha2002facial,he2015multimodal} but these networks were not deep enough or used other feature extraction techniques like PCA or Fisherface. By contrast this paper proposes an end to end system which takes a sequence of frames as input and gives classification labels as output while using deep autoencoders to generate high dimensional spatio-temporal features.

While deep neural nets are notorious for stellar results, training a neural net can be challenging because of huge data requirements. A way around this is to use autoencoders for feature extraction or weights initialization \cite{vincent2008extracting}, followed by fine tuning over a smaller labeled dataset. This issue can also be solved using embeddings in lower dimensional manifold \cite{weston2012deep,kingma2014semi} or pre-train using pseudo labels \cite{lee2013pseudo} thereby requiring fewer number of labeled samples. Approaches based on semi supervised learning have shown to work for smaller labeled datasets \cite{papandreou2015weakly} and techniques using deep neural nets to combine labels and unlabeled data in the same architecture \cite{liu2014facial,kahou2013combining} have emerged victorious. In this paper we propose similar hybrid approaches incorporating deep autoencoders for unlabeled data and additive loss function for the classification tasks.

Introducing invariants in neural networks is an area of active research, some examples include illumination invariant face recognition techniques \cite{mathur2008illumination,li2004illumination} and deep lambertian networks \cite{tang2012deep,jung2015joint}. Our method tries to introduce similar invariants for video neural networks by introducing \textit{temporal} invariants to illumination. While we test our techniques on facial gesture datasets, in principal they can be extended to any neural network taking videos as input.
In \cite{anonymousflorentine}, the author considered velocity changes in videos as well as a semi-supervised learning approach. Here we focus on a different neural network topology and parameter calibration, and report better results on similar databases using new invariant layers.

\section{Method}

\begin{figure*}[t]
\begin{center}
\includegraphics[width=\linewidth]{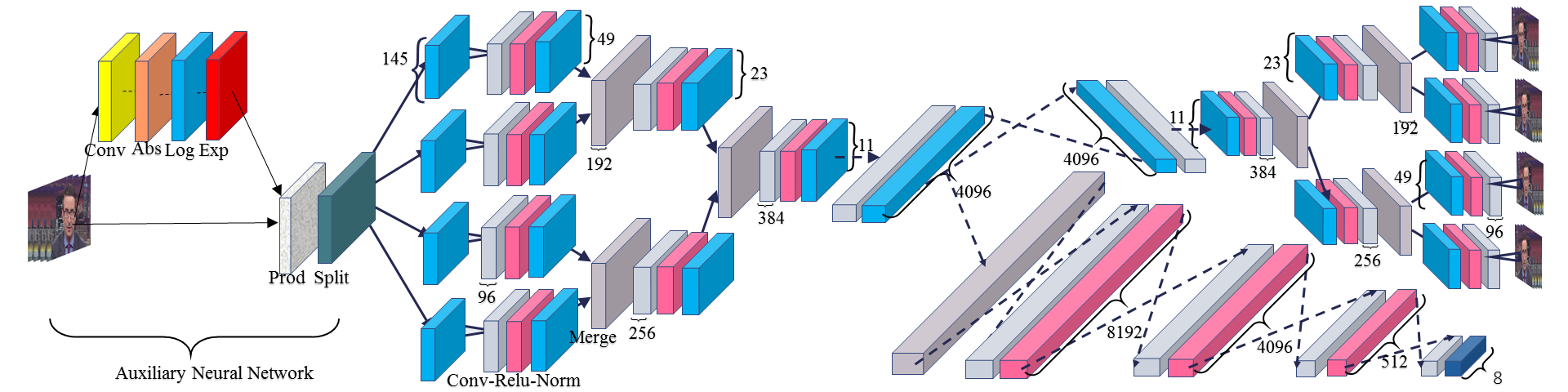}
\end{center}
   \caption{We learn 7 different facial emotions from short (about 1 sec, 25 frames) video clips. The Illumination invariant aspect is achieved by adding an illumination invariant neural network to induce illumination invariance on original input. Our prediction system is based on slow temporal fusion neural network, trained by hybridization of autoencoding a huge collected dataset and a loss prediction on a small set of labeled gestures.  This figure shows the complete architecture combining both illumination invariant and predictor components.}
\label{fig:sigauxillary}
\end{figure*}

\begin{figure}
\begin{tabularx}{\textwidth}{c}
{
\begin{overpic}[width=\linewidth]{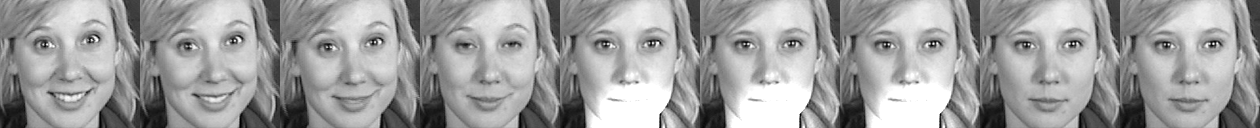}
\put(0,8){\color{white}(a)}
\end{overpic}
} \\
{
\begin{overpic}[width=\linewidth]{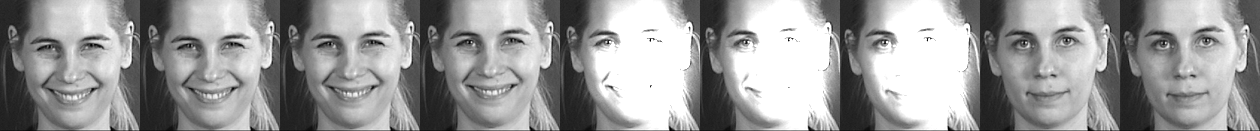}
\put(0,8){\color{white}(b)}
\end{overpic}
} \\
{
\begin{overpic}[width=\linewidth]{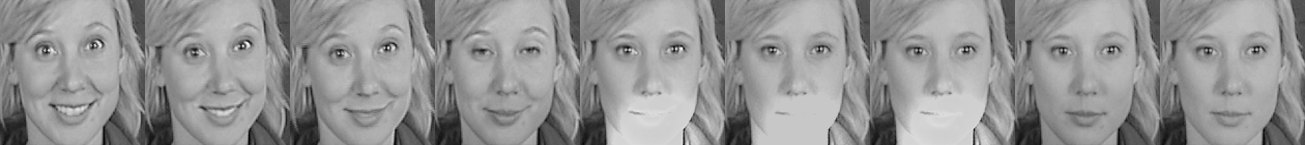}
\put(0,8){\color{white}(c)}
\end{overpic}
} \\
{
\begin{overpic}[width=\linewidth]{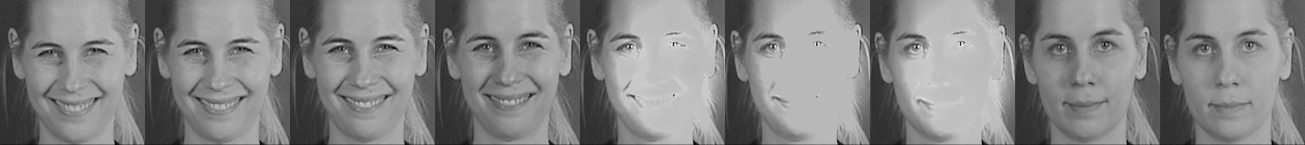}
\put(0,8){\color{white}(d)}
\end{overpic}
}
\end{tabularx}
\caption{Results from autoencoder reconstruction while using scale invariant autoencoder. (a), (b) Input video sequence. (c), (d) Output video sequence from illumination invariant neural network.}
\end{figure}

Our facial expression recognition pipeline comprises of Viola-Jones algorithm \cite{viola2004robust} for face detection followed by a deep convolutional neural network for predicting expressions. The deep convolutional network includes an autoencoder combined with a predictor which relies on the semi-supervised learning paradigm. The autoencoder neural network takes videos containing $9$ frames of size $145 \times 145$ as input and produces $145 \times 145 \times 9$ tensor as output. Predictor neural net sources innermost hidden layer of autoencoder and uses a cascade of fully connected layers accompanied by the softmax layer to classify expressions. Since videos can have different sizes and durations they need to be resized in temporal and spatial domain using standard interpolation techniques. The network topologies and implementation are describe henceforth.

\subsection{Autoencoder}
\label{method:autoencoder}

Stacked autoencoders can be used to convert high dimensional data into lower dimensional space which can be useful for classification, visualization or retrieval \cite{hinton2006reducing}. Since video data is extremely high dimensional we rely on a deep convolutional autoencoder to extract meaningful features from this data by embedding it into $\mathbb{R}^{4096}$. The autoencoder topology is inspired by ImageNet \cite{krizhevsky2012imagenet} and comprises of convolutional layers gradually reducing data dimensionality until we reach a fully connected layer. Central fully connected layers are followed by a cascade of deconvolutional layers which essentially invert the convolutional layers thereby reconstructing the input tensor ($\mathbb{R}^{145 \times 145 \times 9}$). The complete autoencoder architecture can be described in following shorthand $C(96,11,3)-N-C(256,5,2)-N-C(384,3,2)-N-FC(4096)-FC(4096)-DC(96,11,3)-N-DC(256,5,2)-N-DC(384,3,2)$. Here $C(96,11,3)$ is a convolutional layer containing $96$ filters of size $11 \times 11$ in spatial domain and spanning $3$ frames in temporal domain. $N$ stands for local response normalization layers, $DC$ stands for deconvolutional layers and $FC(4096)$ stands for fully connected layers containing $4096$ neurons.

In the same way that spatial convolutions consolidate nearby spatial characteristics of an image, we use the slow fusion model described in \cite{karpathy2014large} to gradually combine temporal features across multiple frames. We implement slow fusion by extending spatial convolution to the temporal domain and adding representation of filter \textit{stride} for both space and time domains. This allows us to control filter size and stride in both temporal and spatial domains leading to a generalized $3\mathbb{D}$ convolution over spatio-temporal input tensor followed by $4\mathbb{D}$ convolutions on intermediate layers. The first convolutional layer sets temporal size and stride as 3 and 2 respectively whereas the subsequent layer has both size and stride of 2 in temporal domain. Finally the third convolutional layer merges temporal information from all frames together, culminating in a lower dimensional vector of size $4096$ at the innermost layer.

Since weight initialization is critical for convergence in a deep autoencoder, we use pre-training for each convolutional layer as we add the layers on. Instead of initializing all weights at once and training from there, we train the first and last layer first, followed by the next convolutional layer and so on. We discuss this in detail in section \ref{experiments_and_results:videoautoencoder}.

\subsection{Semi-Supervised Learner}

\begin{figure*}[t]
\begin{center}
\includegraphics[width=\linewidth]{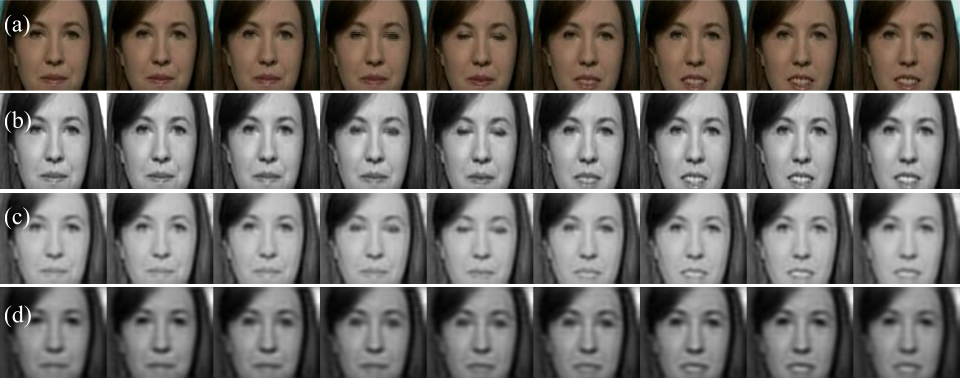}
\end{center}
   \caption{Results from reconstruction using temporal convolutional autoencoder on a face video. (a) Input video sequence. (b) Reconstruction after using 4 convolutional layers. (c) Reconstruction after using 8 layers. (d) Reconstruction after using 12 layers.}
\label{fig:autoencoder_results}
\end{figure*}
\label{datasets}

Our predictor neural net consists of a combination of several convolutional layers followed by multiple fully connected layers ending in a softmax logistic regression layer for prediction. Architecture can be described as $C(96,11,3)-N-C(256,5,2)-N-C(384,3,2)-N-FC(4096)-FC(8192)-FC(4096)-FC(1000)-FC(500)-FC(8)$ using shorthand notation described in section \ref{method:autoencoder}. Notice that our autoencoder architecture is overlaid on top of the predictor architecture by adding deconvolutional layers after the first fully connected layer to create a semi-supervised topology which is capable of training both autoencoder and predictor together (see Figure \ref{fig:sigauxillary}). We use autoencoder to initialize weights for all convolutional layers, all deconvolutional layers and central fully connected layers and we initialize any remaining layers randomly. We use stochastic gradient descent to train weights by combining losses from both predictor and autoencoder while training, this combined loss function for the semi-supervised learner is described in the equation \ref{predictorloss}.

\begin{equation}
\label{predictorloss}
L = - \beta\sum_j{y_j log \left(\frac{e^{o_j}}{\sum_k{e^{o_k}}}\right)} + \alpha||\bar{x}-\bar{x}_o||_2
\end{equation}

Equation \ref{predictorloss} defines semi-supervised learner loss by combining the loss terms from predictor and autoencoder neural networks. Here $y_j$ refers to the input labels to represent each facial expression uniquely while $o_k$ are the outputs from the final layer of predictor neural net. Also $\bar{x}$ is the input tensor ($\in \mathbb{R}^{145 \times 145 \times 9}$) and $\bar{x}_o$ is the corresponding output from autoencoder. Autoencoder loss is the Euclidean loss between input and output tensors given by $||\bar{x}-\bar{x}_o||_2$ whereas $- \sum_j{y_j  log \left(\frac{e^{o_j}}{\sum_k{e^{o_k}}}\right)}$ is the softmax loss from the predictor \cite{bengio2005convex}. Each step of stochastic gradient descent is performed over a batch of 22 inputs and loss is obtained by adding loss terms for the entire batch. At the commencement of training of the predictor layers, we select values of $\beta$ which make softmax loss term an order of magnitude higher than the Euclidean loss term (see equation \ref{predictorloss}). We continue training predictor layers by gradually decreasing loss coefficient $\alpha$ alongside of softmax loss to prevent overfitting of autoencoder. Amalgamation of predictor and autoencoder architectures is depicted in Figure \ref{fig:semisupervisedlearner}.

\subsection{Illumination Invariant Learner}\label{methods:semisupervised}
We introduce scale invariance to pixel intensities by adding additional layers as an illumination invariant neural network in the beginning of semi-supervised learner. The illumination invariant layers include a convolutional layer, an absolute value layer, a reciprocal layer followed by a Hadamard product layer. Scale invariance is achieved by applying element wise multiplication between the output layers of proposed architecture and the original input layer. This normalization can be written as $C(9,1,9)-Abs-Log(\alpha,\beta)-Exp(-\gamma,\delta)-Prod(x_1, x_2)$ (please refer to shorthand notation in section \ref{method:autoencoder}). Here $C(9,1,9)$ refers to the first convolutional layer containing 9 filters with size $1 \times 1$ in spatial domain and a size of $9$ in time domain. $Abs$ is a fixed layer to compute absolute value, $Log(\alpha,\beta)$ layer computes the function $ln(\alpha*x+\beta)$ and $Exp(\gamma,\delta)$ layer gives us $e^{\gamma*x+\delta}$. In the end $Prod(x_1,x_2)$ layer takes two inputs $(x_1,x_2)$ and multiplies the output of exponential layer $(x_2)$ with the original input tensor ($x_1$). If $\bar{F}(\bar{x})$ denotes function emulated by first convolution layer, we can write the transfer function of this sub-net as follows (equation \ref{eq:scaleinvlearner}).

\begin{equation}
\label{eq:scaleinvlearner}
H(\bar{x}) = \bar{x}e^{- \gamma log(\alpha|\bar{F}(\bar{x})| + \beta) + \delta} = \frac{e^{\delta}\bar{x}}{(\alpha|\bar{F}(\bar{x})| + \beta)^\gamma}
\end{equation}

$Log$ and $Exp$ layers are used to generate a reciprocal layer by setting meta-parameters $\gamma$ to $1$ and $\delta$ to zero. We can also \textit{"switch off"}  this sub-net by setting both of these parameters to zero. Transfer function meta parameters $\alpha$ (\textit{scale}) and $\beta$ (\textit{shift}) can be tuned as well for optimal performance. We perform a grid search to find optimal values for these after re-characterizing the transfer function parameters as a global multiplicative factor $\tau$ and a proportion factor $\eta$ (see equation \ref{eq:scaleinvgridsearch}). Table \ref{table:scgridsearch} shows results for various choices of $\alpha$ and $\beta$. We can reformulate equation \ref{eq:scaleinvlearner} as given below:

\begin{equation}
\label{eq:scaleinvgridsearch}
H(\bar{x}) = \frac{e^{0}\bar{x}}{(\alpha|\bar{F}(\bar{x})| + \beta)^1} = \frac{\frac{1}{\beta}\bar{x}}{1 + \frac{\alpha}{\beta}|\bar{F}(\bar{x})|} = \frac{\tau\bar{x}}{1 + \eta|\bar{F}(\bar{x})|}
\end{equation}

The output from scale invariant neural net is a $145 \times 145 \times 9$ tensor which is used as input in the autoencoder and predictor neural networks. The convolution layer can be parametrized using a $9 \times 1 \times 1 \times 9$ tensor and changes during fine tuning while $\alpha$ and $\beta$ are fixed constants greater than zero. In our experiments we initialized convolutional filter of scale invariant sub-net using several approaches, such as partial derivatives, mellin transform, moving average and laplacian kernel and found that it performed best when using neighborhood averaging. Algorithm \ref{bsplinealgo} demonstrates initialization of convolutional layer at the beginning of illumination invariant neural net.

\begin{table*}[]
\centering
\resizebox{\textwidth}{!}{%
\begin{tabular}{|
>{\columncolor[HTML]{CAFEB8}}c |c|
>{\columncolor[HTML]{FDFD96}}c |
>{\columncolor[HTML]{63E9FC}}c |c|
>{\columncolor[HTML]{FDFD96}}c |
>{\columncolor[HTML]{63E9FC}}c |c|
>{\columncolor[HTML]{FDFD96}}c |
>{\columncolor[HTML]{63E9FC}}c |}
\hline
\cellcolor[HTML]{FF9797}$\eta$ $\rightarrow$ & \multicolumn{3}{c|}{\cellcolor[HTML]{FF9797}} & \multicolumn{3}{c|}{\cellcolor[HTML]{FF9797}} & \multicolumn{3}{c|}{\cellcolor[HTML]{FF9797}} \\ \cline{1-1}
$\tau$ $\downarrow$ & \multicolumn{3}{c|}{\multirow{-2}{*}{\cellcolor[HTML]{FF9797}0.1}} & \multicolumn{3}{c|}{\multirow{-2}{*}{\cellcolor[HTML]{FF9797}1}} & \multicolumn{3}{c|}{\multirow{-2}{*}{\cellcolor[HTML]{FF9797}10}} \\ \hline
\cellcolor[HTML]{CAFEB8} & scale ($\eta$/$\tau$) & 0.5 & \cellcolor[HTML]{63E9FC} & scale ($\eta$/$\tau$) & 5 & \cellcolor[HTML]{63E9FC} & scale ($\eta$/$\tau$) & 50 & \cellcolor[HTML]{63E9FC} \\ \cline{2-3} \cline{5-6} \cline{8-9}
\multirow{-2}{*}{\cellcolor[HTML]{CAFEB8}0.2} & shift (1/$\tau$) & 5 & \multirow{-2}{*}{\cellcolor[HTML]{63E9FC}0.486} & shift (1/$\tau$) & 5 & \multirow{-2}{*}{\cellcolor[HTML]{63E9FC}0.472} & shift (1/$\tau$) & 5 & \multirow{-2}{*}{\cellcolor[HTML]{63E9FC}0.243*} \\ \hline
\cellcolor[HTML]{CAFEB8} & scale ($\eta$/$\tau$) & 0.2 & \cellcolor[HTML]{63E9FC} & scale ($\eta$/$\tau$) & 2 & \cellcolor[HTML]{63E9FC} & scale ($\eta$/$\tau$) & 20 & \cellcolor[HTML]{63E9FC} \\ \cline{2-3} \cline{5-6} \cline{8-9}
\multirow{-2}{*}{\cellcolor[HTML]{CAFEB8}0.5} & shift (1/$\tau$) & 2 & \multirow{-2}{*}{\cellcolor[HTML]{63E9FC}0.50} & shift (1/$\tau$) & 2 & \multirow{-2}{*}{\cellcolor[HTML]{63E9FC}0.5135} & shift (1/$\tau$) & 2 & \multirow{-2}{*}{\cellcolor[HTML]{63E9FC}0.45*} \\ \hline
\cellcolor[HTML]{CAFEB8} & scale ($\eta$/$\tau$) & 0.1 & \cellcolor[HTML]{63E9FC} & scale ($\eta$/$\tau$) & 1 & \cellcolor[HTML]{63E9FC} & scale ($\eta$/$\tau$) & 10 & \cellcolor[HTML]{63E9FC} \\ \cline{2-3} \cline{5-6} \cline{8-9}
\multirow{-2}{*}{\cellcolor[HTML]{CAFEB8}1} & shift (1/$\tau$) & 1 & \multirow{-2}{*}{\cellcolor[HTML]{63E9FC}0.499} & shift (1/$\tau$) & 1 & \multirow{-2}{*}{\cellcolor[HTML]{63E9FC}0.51} & shift (1/$\tau$) & 1 & \multirow{-2}{*}{\cellcolor[HTML]{63E9FC}0.47} \\ \hline
\cellcolor[HTML]{CAFEB8} & scale ($\eta$/$\tau$) & 0.02 & \cellcolor[HTML]{63E9FC} & scale ($\eta$/$\tau$) & 0.2 & \cellcolor[HTML]{63E9FC} & scale ($\eta$/$\tau$) & 2 & \cellcolor[HTML]{63E9FC} \\ \cline{2-3} \cline{5-6} \cline{8-9}
\multirow{-2}{*}{\cellcolor[HTML]{CAFEB8}5} & shift (1/$\tau$) & 0.2 & \multirow{-2}{*}{\cellcolor[HTML]{63E9FC}-*} & shift (1/$\tau$) & 0.2 & \multirow{-2}{*}{\cellcolor[HTML]{63E9FC}0.44} & shift (1/$\tau$) & 0.2 & \multirow{-2}{*}{\cellcolor[HTML]{63E9FC}0.50} \\ \hline
\end{tabular}
}
\caption{Accuracies on Florentine (first presented in \protect\cite{anonymousflorentine}) dataset for various values of scale($\alpha$) and shift($\beta$) for illumination invariant neural net. We do a grid search for $\tau$ varying from 0.5 to 5 and $\eta$ varying from 0.1 to 10. Yello columns show corresponding values scale and shift and blue columns show test accuracies. Cells marked with asterisk(*) indicate configurations that did not converge during training.}
\label{table:scgridsearch}
\end{table*}

\begin{algorithm}
\caption{Generate convolution layer for scale invariant sub-net}\label{bsplinealgo}
\algorithmicrequire{ Total number of frames $nFrames$, window size $wSize$} \\
\algorithmicensure{ Caffe Weight Matrix $\mat{W}$}
\begin{algorithmic}[1]
\Function{AutoencoderWeights}{nFrames, wSize}
\State $r \gets (wSize-1)/2$
\State $A \gets zeros(nFrames, nFrames)$
\For{$(i \gets 0;i<nFrames;i++)$}
  \State $n \gets min(i, r)$
  \State $n \gets min(n, nFrames-i)$
    \State $A_{i,i-n:i+n} \gets 1/(2n+1)$
\EndFor
\State $\mat{W} \gets \mat{A}$ \\
\Return $\mat{W}$
\EndFunction
\end{algorithmic}
\end{algorithm}

\section{Datasets and Implementation}


Surprisingly, high quality facial expression datasets are hard to come across. In the same way that majority of facial expression algorithms focus on still images, majority of facial gesture datasets rely on images alone and don't emphasize on complete video clips. For accurate analysis we compare our method against external techniques using 3 different datasets. Each of these datasets have facial video clips varying from neutral face to its peak facial expression. Facial expressions can be naturally occurring \textit{(non-posed)} or artificially enacted \textit{(posed)}, we attempt to classify both using our method and compare our results against published techniques. Here we present the two known datasets from literature along with two additional datasets collected by us. The first dataset was used for unsupervised learning and contains 160 million face  images combined into 6.5 million short (25 frames) clips. The second dataset contains 2777 video clips which are labeled for seven basic emotions.

\subsection{Autoencoder dataset}

\begin{table}
\begin{center}
{\renewcommand{\arraystretch}{1.2}
\begin{tabularx}{0.95\linewidth}{|l|>{\centering\arraybackslash}X|>{\centering\arraybackslash}X|>{\centering\arraybackslash}X|}
\hline
Emotion & \cellcolor{hcolor!120} Posed & \cellcolor{hcolor!120} Non-Posed & \cellcolor{hcolor!120} Cumulative \\
\hline
Anger & 132 & 318 & 450 \\
\tdash
Sadness & 118 & 148 & 266 \\
\tdash
Contempt & 153 & 301 & 454 \\
\tdash
Fear & 137 & 96 & 233 \\
\tdash
Surprise & 188 & 232 & 420 \\
\tdash
Joy & 172 & 503 & 675 \\
\tdash
Disgust & 132 & 147 & 279 \\
\hline
Total & \cellcolor{cmcolor!120} 1032 & \cellcolor{cmcolor!120} 1745 & \cellcolor{cmcolor!120} 2777 \\
\hline
\end{tabularx}
}
\end{center}
\caption{Data distribution for \dbname dataset for various emotions. Posed clips refer to the artificially generated clips, while non-posed refer to those captured using the stimulus activation procedure.}
\label{table:Florentinedistribution}
\end{table}

Training the unsupervised component of our neural net required a large amount of data to ensure that the deep features were general enough to represent any face expression. We trained the deep convolutional autoencoder using a massive collection of unlabeled data points comprising of 6.5 million video clips with 25 image frames per clip. The clips were generated by running Viola-Jones face algorithm to detect and isolate face bounding boxes on public domain videos. We further enhanced the data quality by removing any clips which showed high variation in-between consecutive frames. This eliminated video clips containing accidental appearance of occlusions, rapid facial motions or sudden appearance of another face.

As an additional step we obtained the facial pose information by using active appearance models and generating facial landmarks \cite{asthana2014incremental}. We fitted the facial landmarks to a $3D$ deformable model and restricted our dataset to clips containing less than 30 degrees of yaw, pitch or roll, thereby eliminating faces looking sideways. For data generation, we relied on daily feeds from news sources such as CNN, MSNBC, FOX and CSPAN. Collection of this dataset required development of an automated system to mine video clips, segment faces and filter meaningful data and it took us more than 6 months to collect the entire dataset. To our knowledge this is the largest dataset containing facial video clips and we plan to share it with scientific community by making it public.

\subsection{Cohn Kanade Dataset}

The Cohn Kanade dataset \cite{lucey2010extended} is one of the oldest and well known dataset containing facial expression video clips. It contains a total of 593 video clip sequences from which 327 clips are labeled for seven basic emotions (most of these are \textit{posed}). Clips contain the frontal view of face performing facial gesture varying from neutral expression to maximum intensity of emotion. While the dataset contains a lot of natural smile expressions it lacks diversity of induced samples for other facial expressions.


\begin{table*}
\small
\tabcolsep=0.15cm
\begin{tabularx}{\textwidth}{l l} {
\begin{tabularx}{0.5\textwidth}{l!{\color{white}\vrule}c!{\color{white}\vrule}c!{\color{white}\vrule}c!{\color{white}\vrule}c!{\color{white}\vrule}c!{\color{white}\vrule}c!{\color{white}\vrule}c}
\multicolumn{8}{c}{\textbf{Confusion matrix using our methods on \dbname Dataset}}\\
\multicolumn{1}{c}{} & \multicolumn{1}{c}{\textit{Anger}} & \multicolumn{1}{c}{\textit{Contempt}} & \multicolumn{1}{c}{\textit{Happy}} & \multicolumn{1}{c}{\textit{Disgust}} & \multicolumn{1}{c}{\textit{Fear}} & \multicolumn{1}{c}{\textit{Sadness}}& \multicolumn{1}{c}{\textit{Surprise}}\\
 \textit{Anger} & 0.54  \cellcolor{fmcolor!80}
& 0.13  \cellcolor{cmcolor!3}  & 0.01  \cellcolor{cmcolor!0}  & 0.12  \cellcolor{cmcolor!2}  & 0.03  \cellcolor{cmcolor!0}  & 0.12  \cellcolor{cmcolor!1}  & 0.04  \cellcolor{cmcolor!0}  \\
 \textit{Contempt} & 0.07  \cellcolor{cmcolor!0}  & 0.47  \cellcolor{cmcolor!80}
& 0.14  \cellcolor{cmcolor!3}  & 0.06  \cellcolor{cmcolor!0}  & 0.04  \cellcolor{cmcolor!0}  & 0.14  \cellcolor{cmcolor!4}  & 0.07  \cellcolor{cmcolor!0}  \\
 \textit{Happy} & 0.01  \cellcolor{cmcolor!0}  & 0.15  \cellcolor{cmcolor!4}  & 0.78  \cellcolor{fmcolor!80}
& 0.04  \cellcolor{cmcolor!0}  & 0.00  \cellcolor{cmcolor!0}  & 0.00  \cellcolor{cmcolor!0}  & 0.00  \cellcolor{cmcolor!0}  \\
 \textit{Disgust} & 0.13  \cellcolor{cmcolor!2}  & 0.21  \cellcolor{cmcolor!11}  & 0.15  \cellcolor{cmcolor!5}  & 0.31  \cellcolor{cmcolor!80}
& 0.07  \cellcolor{cmcolor!0}  & 0.05  \cellcolor{cmcolor!0}  & 0.08  \cellcolor{cmcolor!0}  \\
 \textit{Fear} & 0.04  \cellcolor{cmcolor!0}  & 0.14  \cellcolor{cmcolor!4}  & 0.08  \cellcolor{cmcolor!0}  & 0.08  \cellcolor{cmcolor!0}  & 0.28  \cellcolor{cmcolor!80}
& 0.04  \cellcolor{cmcolor!0}  & 0.32  \cellcolor{cmcolor!22}  \\
 \textit{Sadness} & 0.18  \cellcolor{cmcolor!8}  & 0.22  \cellcolor{cmcolor!11}  & 0.07  \cellcolor{cmcolor!0}  & 0.10  \cellcolor{cmcolor!0}  & 0.02  \cellcolor{cmcolor!0}  & 0.27  \cellcolor{cmcolor!80}
& 0.13  \cellcolor{cmcolor!3}  \\
 \textit{Surprise} & 0.04  \cellcolor{cmcolor!0}  & 0.07  \cellcolor{cmcolor!0}  & 0.07  \cellcolor{cmcolor!0}  & 0.05  \cellcolor{cmcolor!0}  & 0.21  \cellcolor{cmcolor!10}  & 0.05  \cellcolor{cmcolor!0}  & 0.51  \cellcolor{fmcolor!80}
\\
\multicolumn{8}{c}{}\\
\end{tabularx}
} &
{
\begin{tabularx}{0.5\textwidth}{l c c c c c c c }
\multicolumn{8}{c}{\textbf{Confusion matrix using external methods on \dbname dataset}}\\
\multicolumn{1}{c}{} & \multicolumn{1}{c}{\textit{Anger}} & \multicolumn{1}{c}{\textit{Contempt}} & \multicolumn{1}{c}{\textit{Happy}} & \multicolumn{1}{c}{\textit{Disgust}} & \multicolumn{1}{c}{\textit{Fear}} & \multicolumn{1}{c}{\textit{Sadness}}& \multicolumn{1}{c}{\textit{Surprise}}\\
\hhline{~~~~~~~~}
 \textit{Anger} & 0.61 \cellcolor{cmcolor!80} & 0.08 \cellcolor{cmcolor!0} & 0.06 \cellcolor{cmcolor!0} & 0.07 \cellcolor{cmcolor!0} & 0.04 \cellcolor{cmcolor!0} & 0.06 \cellcolor{cmcolor!0} & 0.08 \cellcolor{cmcolor!0} \\
\hhline{~~~~~~~~}
 \textit{Contempt} & 0.09 \cellcolor{cmcolor!0} & 0.44 \cellcolor{fmcolor!80} & 0.27 \cellcolor{cmcolor!17} & 0.06 \cellcolor{cmcolor!0} & 0.02 \cellcolor{cmcolor!0} & 0.06 \cellcolor{cmcolor!0} & 0.05 \cellcolor{cmcolor!0} \\
\hhline{~~~~~~~~}
 \textit{Happy} & 0.01 \cellcolor{cmcolor!0} & 0.07 \cellcolor{cmcolor!0} & 0.85 \cellcolor{cmcolor!80} & 0.01 \cellcolor{cmcolor!0} & 0 \cellcolor{cmcolor!0} & 0.01 \cellcolor{cmcolor!0} & 0.06 \cellcolor{cmcolor!0} \\
\hhline{~~~~~~~~}
 \textit{Disgust} & 0.19 \cellcolor{cmcolor!8} & 0.14 \cellcolor{cmcolor!4} & 0.25 \cellcolor{cmcolor!30} & 0.22 \cellcolor{fmcolor!80} & 0.01 \cellcolor{cmcolor!0} & 0.02 \cellcolor{cmcolor!0} & 0.16 \cellcolor{cmcolor!6} \\
\hhline{~~~~~~~~}
 \textit{Fear} & 0.21 \cellcolor{cmcolor!10} & 0.09 \cellcolor{cmcolor!0} & 0.06 \cellcolor{cmcolor!0} & 0.07 \cellcolor{cmcolor!0} & 0.12 \cellcolor{fmcolor!80} & 0.06 \cellcolor{cmcolor!0} & 0.39 \cellcolor{cmcolor!80} \\
\hhline{~~~~~~~~}
 \textit{Sadness} & 0.26 \cellcolor{cmcolor!15} & 0.20 \cellcolor{cmcolor!9} & 0.19 \cellcolor{cmcolor!8} & 0.02 \cellcolor{cmcolor!0} & 0.05 \cellcolor{cmcolor!0} & 0.13 \cellcolor{fmcolor!80} & 0.16 \cellcolor{cmcolor!6} \\
\hhline{~~~~~~~~}
 \textit{Surprise} & 0.15 \cellcolor{cmcolor!4} & 0.03 \cellcolor{cmcolor!0} & 0.09 \cellcolor{cmcolor!0} & 0.01 \cellcolor{cmcolor!0} & 0.06 \cellcolor{cmcolor!0} & 0.05 \cellcolor{cmcolor!0} & 0.61 \cellcolor{cmcolor!80} \\
\hhline{~~~~~~~~}
\multicolumn{8}{c}{}\\
\end{tabularx}
} \\
{
\begin{tabularx}{0.5\textwidth}{l c c c c c c c}
\multicolumn{8}{c}{\textbf{Confusion matrix using our methods on Cohn-Kanade}}\\
\multicolumn{1}{c}{} & \multicolumn{1}{c}{\textit{\textit{Anger}}} & \multicolumn{1}{c}{\textit{Contempt}} & \multicolumn{1}{c}{\textit{Happy}} & \multicolumn{1}{c}{\textit{Disgust}} & \multicolumn{1}{c}{\textit{Fear}} & \multicolumn{1}{c}{\textit{Sadness}}& \multicolumn{1}{c}{\textit{Surprise}}\\
\hhline{~~~~~~~~}
  \textit{Anger} & 0.92  \cellcolor{cmcolor!100}  & 0.08  \cellcolor{cmcolor!0}  & 0  \cellcolor{cmcolor!0}  & 0  \cellcolor{cmcolor!0}  & 0  \cellcolor{cmcolor!0}  & 0  \cellcolor{cmcolor!0}  & 0  \cellcolor{cmcolor!0}  \\
\hhline{~~~~~~~~}
 \textit{Contempt} & 0  \cellcolor{cmcolor!0}  & 0.80  \cellcolor{fmcolor!100}  & 0  \cellcolor{cmcolor!0}  & 0  \cellcolor{cmcolor!0}  & 0  \cellcolor{cmcolor!0}  & 0  \cellcolor{cmcolor!0}  & 0.20  \cellcolor{cmcolor!9}  \\
\hhline{~~~~~~~~}
 \textit{Happy} & 0  \cellcolor{cmcolor!0}  & 0  \cellcolor{cmcolor!0}  & 1.00  \cellcolor{cmcolor!100}  & 0  \cellcolor{cmcolor!0}  & 0  \cellcolor{cmcolor!0}  & 0  \cellcolor{cmcolor!0}  & 0  \cellcolor{cmcolor!0}  \\
\hhline{~~~~~~~~}
 \textit{Disgust} & 0.06  \cellcolor{cmcolor!0}  & 0  \cellcolor{cmcolor!0}  & 0  \cellcolor{cmcolor!0}  & 0.94  \cellcolor{cmcolor!100}  & 0  \cellcolor{cmcolor!0}  & 0  \cellcolor{cmcolor!0}  & 0  \cellcolor{cmcolor!0}  \\
\hhline{~~~~~~~~}
 \textit{Fear} & 0  \cellcolor{cmcolor!0}  & 0  \cellcolor{cmcolor!0}  & 0.14  \cellcolor{cmcolor!4}  & 0  \cellcolor{fmcolor!0}  & 0.71  \cellcolor{fmcolor!60}  & 0  \cellcolor{cmcolor!0}  & 0.14  \cellcolor{cmcolor!4}  \\
\hhline{~~~~~~~~}
 \textit{Sadness} & 0.38  \cellcolor{cmcolor!27}  & 0  \cellcolor{cmcolor!0}  & 0  \cellcolor{cmcolor!0}  & 0  \cellcolor{cmcolor!0}  & 0  \cellcolor{cmcolor!0}  & 0.50  \cellcolor{cmcolor!79}  & 0.12  \cellcolor{cmcolor!2}  \\
\hhline{~~~~~~~~}
 \textit{Surprise} & 0  \cellcolor{cmcolor!0}  & 0  \cellcolor{cmcolor!0}  & 0  \cellcolor{cmcolor!0}  & 0  \cellcolor{cmcolor!0}  & 0  \cellcolor{cmcolor!0}  & 0  \cellcolor{cmcolor!0}  & 1.00  \cellcolor{cmcolor!100}  \\
\hhline{~~~~~~~~}
\multicolumn{8}{c}{}\\
\end{tabularx}
} &
{
\begin{tabularx}{0.5\textwidth}{l c c c c c c c }
\multicolumn{8}{c}{\textbf{Confusion matrix using external methods on Cohn-Kanade}}\\
\multicolumn{1}{c}{} & \multicolumn{1}{c}{\textit{Anger}} & \multicolumn{1}{c}{\textit{Contempt}} & \multicolumn{1}{c}{\textit{Happy}} & \multicolumn{1}{c}{\textit{Disgust}} & \multicolumn{1}{c}{\textit{Fear}} & \multicolumn{1}{c}{\textit{Sadness}} & \multicolumn{1}{c}{\textit{Surprise}}\\
\hhline{~~~~~~~~}
 \textit{Anger} & 0.73 \cellcolor{fmcolor!100} & 0 \cellcolor{cmcolor!0} & 0.07 \cellcolor{cmcolor!0} & 0 \cellcolor{cmcolor!0} & 0 \cellcolor{cmcolor!0} & 0.20 \cellcolor{cmcolor!9} & 0 \cellcolor{cmcolor!0} \\
\hhline{~~~~~~~~}
 \textit{Contempt} & 0 \cellcolor{cmcolor!0} & 0.86 \cellcolor{cmcolor!100} & 0 \cellcolor{cmcolor!0} & 0 \cellcolor{cmcolor!0} & 0 \cellcolor{cmcolor!0} & 0.14 \cellcolor{cmcolor!4} & 0 \cellcolor{cmcolor!0} \\
\hhline{~~~~~~~~}
 \textit{Happy} & 0 \cellcolor{cmcolor!0} & 0 \cellcolor{cmcolor!0} & 0.95 \cellcolor{fmcolor!100} & 0 \cellcolor{cmcolor!0} & 0.05 \cellcolor{cmcolor!0} & 0 \cellcolor{cmcolor!0} & 0 \cellcolor{cmcolor!0} \\
\hhline{~~~~~~~~}
 \textit{Disgust} & 0.25 \cellcolor{cmcolor!14} & 0.12 \cellcolor{cmcolor!2} & 0 \cellcolor{cmcolor!0} & 0.38 \cellcolor{fmcolor!80} & 0 \cellcolor{cmcolor!0} & 0.12 \cellcolor{cmcolor!2} & 0.12 \cellcolor{cmcolor!2} \\
\hhline{~~~~~~~~}
 \textit{Fear} & 0 \cellcolor{cmcolor!0} & 0 \cellcolor{cmcolor!0} & 0 \cellcolor{cmcolor!0} & 0 \cellcolor{cmcolor!0} & 1.00 \cellcolor{cmcolor!100} & 0 \cellcolor{cmcolor!0} & 0 \cellcolor{cmcolor!0} \\
\hhline{~~~~~~~~}
 \textit{Sadness} & 0.33 \cellcolor{cmcolor!23} & 0 \cellcolor{cmcolor!0} & 0 \cellcolor{cmcolor!0} & 0.11 \cellcolor{cmcolor!1} & 0 \cellcolor{cmcolor!0} & 0.44 \cellcolor{fmcolor!60} & 0.11 \cellcolor{cmcolor!1} \\
\hhline{~~~~~~~~}
 \textit{Surprise} & 0 \cellcolor{cmcolor!0} & 0 \cellcolor{cmcolor!0} & 0 \cellcolor{cmcolor!0} & 0.05 \cellcolor{cmcolor!0} & 0 \cellcolor{cmcolor!0} & 0 \cellcolor{cmcolor!0} & 0.95 \cellcolor{fmcolor!100} \\
\hhline{~~~~~~~~}
\multicolumn{8}{c}{}\\
\end{tabularx}
} \\
\end{tabularx}
\caption{Confusion matrices over test results for Cohn Kanade and Florentine datasets using our methods and best performing external method which uses \textit{Expressionlets} for CKPlus \protect\cite{liu2014learning} and \textit{Covariance Riemann kernel} for Florentine \protect\cite{liu2014combining}.  On the left we show results for the proposed illumination invariant semi-supervised approach across various facial expressions, while on the right we present confusion matrix from external methods. Highest accuracy in each category is marked using blue color. For CKPlus we outperform competing method in 5 verticals by getting 100\% accuracy on happiness, 100\% on surprise, 94\% on disgust, 92\% in anger and 50\% in sadness. For both methods misclassification occur when emotions like sadness get recognized as anger and \textit{vice-versa}.}
\label{table:cfckplus}
\end{table*}

\subsection{MMI Dataset}

MMI facial expression dataset \cite{pantic2005web} involves an ongoing effort for representing both enacted and induced facial expressions. The dataset comprises of 2894 video samples out of which around 200 video clips are labeled for six basic emotions. The clips contain faces going from blank expression to the peak emotion and then back to neutral facial gesture. MMI which originally contained only posed facial expressions, was recently extended to include natural versions of happiness, disgust and surprise \cite{valstar2010induced}.

\subsection{\dbname dataset}

We developed specialized video recording and annotation tools to collect and label facial gestures (first presented in \cite{anonymousflorentine}). The application was developed in Python programming language and we used well known libraries such as OpenCV for video capture and annotation. The database contains facial clips from 160 subjects (both male and female), where gestures were artificially generated according to a specific request, or genuinely given due to a shown stimulus. We captured 1032 clips for posed expressions and 1745 clips for induced facial expressions amounting to a total of 2777 video clips. Genuine facial expressions were induced in subjects using visual stimuli, i.e. videos selected randomly from a bank of Youtube videos to generate a specific emotion. Please refer to Table \ref{table:Florentinedistribution} to see the distribution of database, where posed clips refers to the artificially generated expressions and non-posed refers to the stimulus activation procedure.

\section{Experiments and results}

\subsection{Video autoencoder}\label{experiments_and_results:videoautoencoder}
Since deep autoencoders can show slow convergence when trained from randomly initialized weights \cite{hinton2006reducing}, we used contrastive divergence minimization to train stacked autoencoder layers \textit{iteratively} \cite{carreira2005contrastive}. Initially, we pre-trained the beginning and end convolutional layers by creating an intermediate neural network $(C(96,11,3)-N-C(256,5,2)-N-DC(256,5,2)-N-DC(384,3,2))$ and training it on facial video clips. Inner layers were trained successively by adding them to the intermediate neural network and keeping pre-trained layers fixed until the convergence of weights. To yield best results, we also fine tuned the entire network at the end of each iteration. This process was repeated until the required number of layers had been added and final architecture was achieved. Training of the entire autoencoder typically required 3 days and a million data inputs.

\begin{table*}[]
\centering
\subfloat[Comparison of results using Illumination Invariant Techniques for datasets under standard conditions.]{\begin{tabular}{|l|c|c|c|}
\hline
                        & \multicolumn{1}{l|}{\cellcolor[HTML]{\colora}\textit{\textbf{MMI}}} & \multicolumn{1}{l|}{\cellcolor[HTML]{\colora}\textit{\textbf{CKPlus}}} & \multicolumn{1}{l|}{\cellcolor[HTML]{\colora}\textit{\textbf{Florentine}}} \\ \hline
Main Gaussian Riemann   & 40.9                                                               & 67                                                                    & 46.92                                                                     \\ \hline
Main Grassman           & 9.09                                                               & 17.9                                                                  & 44.99                                                                     \\ \hline
Main Covariance Reimann & 40.9                                                               & 79                                                                    & \cellcolor[HTML]{\colorb}51.05                                             \\ \hline
Expressionlets          & \cellcolor[HTML]{\colorb}52.91                                      & \cellcolor[HTML]{\colorb}82.7                                          & 48.6                                                                      \\ \hline
Semi-Supervised Learner                 & 59.01                                                              & 87.36                                                                 & 51.11                                                                     \\ \hline
Scale Invariant Learner     & \cellcolor[HTML]{\colorc}65.57                                      & \cellcolor[HTML]{\colorc}90.52                                         & \cellcolor[HTML]{\colorc}51.35                                             \\ \hline
\end{tabular}}
\quad
\subfloat[Comparison of results using Illumination Invariant Techniques for datasets under changing lighting.]{\begin{tabular}{|l|c|c|c|}
\hline
                        & \multicolumn{1}{l|}{\cellcolor[HTML]{\colora}\textit{\textbf{MMI}}} & \multicolumn{1}{l|}{\cellcolor[HTML]{\colora}\textit{\textbf{CKPlus}}} & \multicolumn{1}{l|}{\cellcolor[HTML]{\colora}\textit{\textbf{Florentine}}} \\ \hline
Main Gaussian Riemann   & 39.39                                                              & 65                                                                    & 43.9                                                                      \\ \hline
Main Grassman           & 9.09                                                               & 11                                                                    & 43.91                                                                     \\ \hline
Main Covariance Reimann & 36.36                                                              & 65                                                                    & \cellcolor[HTML]{\colorb}46.44                                                                     \\ \hline
Expressionlets          & \cellcolor[HTML]{\colorb}55.38                                      & \cellcolor[HTML]{\colorb}70                                            & 46.02                                             \\ \hline
Semi-Supervised Learner                  & \multicolumn{1}{l|}{55.7}                                          & 68.42                                          &38.66                                                \\ \hline
Scale Invariant Learner       & \multicolumn{1}{l|}{\cellcolor[HTML]{\colorc}59.03}                 & \cellcolor[HTML]{\colorc}73.68                    & \cellcolor[HTML]{\colorc}48                           \\ \hline
\end{tabular}}
\caption{Comparison of results from various techniques on CKPlus, MMI and \dbname datasets. The dataset was divided into 3 parts test, train and val randomly. Training set was 50\%, test and validation were 30\% and 20\% respectively. The table on the left shows results on original data while the one on right shows results after we added illumination changes. Our method consistently won for both small and large datasets (winning method is shown in blue and the leading method is showed using yellow).}
\label{table:predictor_results}
\end{table*}

Our neural network was implemented using the \textit{Caffe} framework \cite{jia2014caffe} and trained using NVIDIA Tesla K40 GPUs. The trained weights used to initialize next phase were stored as Caffe \textit{model} files and each intermediate neural network was implemented as a separate \textit{prototxt} file. Weights were shared using shared parameter feature and transferred across neural networks using the resume functionality provided in \textit{Caffe}. Our deep autoencoder took $145 \times 145 \times 9$ clips as input, the spatial resolution was achieved by down-sampling all clips to a fixed size using bi-cubic interpolation. $9$ frames were obtained by extracting every third frame from video clips. All videos were converted into $1305 \times 145$ image clips containing consecutive input frames placed horizontally and we used the  \textit{Caffe "imagedata", "split"} and \textit{"concat"} layers to isolate individual frames for autoencoder input and output.

Please see Figure \ref{fig:autoencoder_results} to visualize results obtained from intermediate autoencoders using different number of layers.

\subsection{Semi-Supervised predictor}\label{experiments_and_results:semisupervised}

We created a semi-supervised predictor by adding a deep neural network after the innermost fully connected layer of our autoencoder. The architecture of predictor neural net can be written as $FC(8192)-FC(4096)-FC(1000)-FC(500)-FC(8)$. The complete semi-supervised neural network contains an autoencoder and a predictor that share neural links and can be trained on the same input simultaneously. Weights from autoencoder training were used to initialize weights of semi-supervised predictor which were later fine tuned using labeled inputs from datasets described in section \ref{datasets}. The weights from this step are used for initialization of our scale-invariant predictor which we describe next.

\subsection{Illumination-Invariant Semi-Supervised predictor}
Our scale-invariant neural network prefixes semi-supervised learner with an axillary neural net to induce scale invariance (see \ref{methods:semisupervised}). We test our method on three datasets (MMI, CK and \dbname) by randomly dividing each of them into non-intersecting train, test and validation subsets. Our training dataset contains  50\% inputs while testing and validation datasets contain 30\% and 20\% of inputs. After the split we increase the size of training dataset by adding rotation, translation or flipping the image.

For quantitative analysis we compare our results against expression-lets base approaches \cite{liu2014learning} and multiple kernel methods \cite{liu2014combining}. We utilize sources downloaded from \textit{Visual data transforming and taking in Resources} \cite{vipl} as a reference to contrast with our strategies. For reasonable comparison we use same partitioning techniques while comparing our techniques with external methods. While we cannot compare against methods such as \cite{liu2014deeply} because of absence of publicly available code our method still wins on MMI dataset.

We test our method with and without varying illumination on external datasets, results of our findings can be summarized in Table \ref{table:predictor_results}. Please see tables \ref{table:cfckplus} for confusion matrices demonstrating results for each expression. We outperform all external methods on datasets in almost all cases. Our method also shows large margin of improvement over plain semi-supervised approaches. Both autoencoder and predictor network topologies are implemented as \textit{Caffe} \textit{prototxt} files \cite{jia2014caffe} and they will be made available for public usage.

\section{Discussions and future work}

In this paper we introduce a framework for facial gesture recognition which combines semi-supervised learning approaches with carefully designed  neural network topologies. We demonstrate how to induce illumination invariance by including specialized layers and use spatio-temporal convolutions to extract features from multiple image frames. Currently, our system relies on utilization of Viola-Jones to distinguish and segment out the faces and is limited to analyzing only the front facing views. Emotion recognition in the wild still remains an elusive problem with low reported accuracies which we hope will be addresses in future work.

In this work we only considered video frames but other, richer, modalities could be taken into account. Sound, for example, has a direct influence on the emotional status and can improve our current system. Higher refresh rates, multi-resolution in space and time, or interactions between our subjects are just few of many possibilities which can to enrich our data and can
lead to better classification or inference.



Deep neural networks have proven to be extremely effective in solving computer vision problems even though training them at large scale continues to be both CPU and memory intensive. Our system tries to make best use of resources available and further improvements in hardware and software can help us build even larger and deeper neural networks while enabling us to train and test them on portable devices. Over here, we introduce a new layer which creates illumination invariance \textit{adaptively} and can be fine tuned to get best results. In this work, we emphasize on scale invariance for illumination, in future we hope to explore induction of other invariants, which continues to be an area of rapid research in neural networks.

Another approach to induce scale invariance can involve using standardized Local Response Normalization (LRN) based layers in the neural network right after the first input layer. This approach is similar to pre-normalizing the data before testing. We compare our method to this approach as well and found that adaptive normalization performed better than plain LRN based learner. Our results are summarized in Table \ref{table:lrn_results}.

\begin{table}
\begin{tabular}{|l|c|c|c|}
\hline
& \multicolumn{1}{l|}{\cellcolor[HTML]{\colora}\textit{\textbf{MMI}}} & \multicolumn{1}{l|}{\cellcolor[HTML]{\colora}\textit{\textbf{CKPlus}}} & \multicolumn{1}{l|}{\cellcolor[HTML]{\colora}\textit{\textbf{Florentine}}} \\ \hline
Semi-Supervised Learner                  & \multicolumn{1}{l|}{55.7}                                          & 68.42                                          & \cellcolor[HTML]{\colorb}38.66                                                \\ \hline
LRN @(0.5)                  & \multicolumn{1}{l|}{54.09}                                          & \cellcolor[HTML]{\colorb}69.47                                         &35                                                \\ \hline
LRN @(0.75)                  & \multicolumn{1}{l|}{\cellcolor[HTML]{\colorb}55.73}                                          & \cellcolor[HTML]{\colorb}69.47                                         &35.84                                                \\ \hline
Scale Invariant Learner       & \multicolumn{1}{l|}{\cellcolor[HTML]{\colorc}59.03}                 & \cellcolor[HTML]{\colorc}73.68                    & \cellcolor[HTML]{\colorc}48                           \\ \hline
\end{tabular}
\caption{Comparison of illumination invariant learner to plain semi supervised learner with Local Response Normalization layers in the beginning. We try changing coeffecients of Local Response Normalization and got good results when setting $\beta$ at $0.75$. Our method continued to win for both small and large datasets (winning method is shown in blue and the leading method is showed using yellow).}
\label{table:lrn_results}
\end{table}

\subsection{Limitations}
In this section we explore limitations of our system and discuss where our system may fail or be of less value. One of our greatest limitations is that the system was built and tested using only frontal perspectives thereby imposing a constraint on the input facial orientations. Further the pipeline takes a fixed number of video frames as input which imposes a restriction on minimum number of frames required for recognition. We restrict individual frames to a fixed size of $140 \times 140$ and higher resolution frames need to be resized which may lead to information loss. Both spatio and temporal size constraints can be improved by increasing neural network size at the cost of computing resources.

Learning for deep neural networks can be extremely computationally intensive and can impose massive constraints on systemic space-time complexity. Our system is no different and requires specialized hardware (NVIDIA Tesla\texttrademark or K40\texttrademark Grid GPUs) with a minimum of 9 GB of VRAM on the graphics card for lowest of batch sizes. Deep autoencoders can be data intensive and require millions of unlabeled samples to train. Further the stacked autoencoder we train takes over 3 days to train requiring an additional day to fine tune predictor weights for larger labeled datasets. Even though the system supports 7 emotions and 1 neutral face state, it was not trained to detect neutral emotions - a constraint which can be fixed by adding more labeled data for neutral facial gestures. The pipeline only recognizes 7 facial emotions but recent research shows that there is a much wider range of emotions. Even though neural networks win in a lot of scenarios, a lot more research needs to be done to understand exactly how and why they work.

\section{Conclusions}

This paper uses semi-supervised paradigms in convolutional neural nets for classification of facial gestures in video sequences. Our topologies are trained on millions of facial video clips and use spatio-temporal convolutions to extract transient features in videos. We developed a new scale-invariant sub-net which showed superior results for gesture recognition under variable lighting conditions. We demonstrate effectiveness of our approach on both publicly available datasets and samples collected by us.

\bibliographystyle{acmsiggraph}
\bibliography{template}
\end{document}